\documentclass[sigconf,9pt]{acmart}
\usepackage{algorithm}
\usepackage{algpseudocode}
\algtext*{EndWhile}
\algtext*{EndIf}
\algtext*{EndFor}
\algtext*{EndProcedure}
\usepackage{amsmath}
\usepackage{amssymb}
\usepackage{array}
\usepackage{booktabs}
\usepackage{color}
\usepackage{comment}
\usepackage{cuted}
\usepackage[inline]{enumitem}
\usepackage{epsfig}
\usepackage{etoolbox}
\usepackage{fancybox}
\usepackage{float}
\usepackage{graphicx}
\usepackage{listings}
\usepackage{lipsum}
\usepackage{multirow}
\usepackage{multicol}
\usepackage{makecell}
\usepackage{placeins}
\usepackage{rotating}
\usepackage{setspace}
\usepackage[hang, tight]{subfigure}
\usepackage{threeparttable}
\usepackage{url}
\usepackage{verbatim}
\usepackage{wrapfig}
\usepackage[dvipsnames]{xcolor}

\newbool{inccomment}
\setbool{inccomment}{true}
\graphicspath{{_fig/}}

\algrenewcommand\algorithmicindent{1.0em}%

\newcommand{\XC}[1]{\ifbool{inccomment}{{\color{magenta}YC\@: #1}}{}}
\newcommand{\ML}[1]{\ifbool{inccomment}{{\color{blue}XC\@: #1}}{}}
\newcommand{\TD}[1]{\ifbool{inccomment}{{\color{orange}#1}}{}}
\newcommand{\FN}[1]{\ifbool{inccomment}{{\color{OliveGreen}#1}}{}}

\newcommand{\ct}{\ifbool{inccomment}{{\color{magenta}$[$C$]$}}}
\newcommand{\llnn}{\ifbool{inccomment}{{\color{magenta}\\=============================================\\}}}

\newcommand{\ours}{\textit{MIRAGE}}

\copyrightyear{2026}
\acmYear{2026}
\setcopyright{cc}
\setcctype{by}
\acmConference[DAC '26]{63rd ACM/IEEE Design Automation Conference}{July 26--29, 2026}{Long Beach, CA, USA}
\acmBooktitle{63rd ACM/IEEE Design Automation Conference (DAC '26), July 26--29, 2026, Long Beach, CA, USA}
\acmDOI{10.1145/3770743.3804297}
\acmISBN{979-8-4007-2254-7/2026/07}

\title{\ours: Runtime Scheduling for Multi-Vector Image Retrieval with Hierarchical Decomposition}
\author{
Maoliang Li$^{1,*}$,
Ke Li$^{2,*}$,
Yaoyang Liu$^{3}$,
Jiayu Chen$^{1}$, 
Zihao Zheng$^{1}$, \\
Yinjun Wu$^{1}$,
Chenchen Liu$^{4}$,
Xiang Chen$^{1,\dagger}$
}

\affiliation{%
  \institution{\small
    $^{1}$School of Computer Science, Peking University\hspace{0.3em}
    $^{2}$School of Electronics Engineering and Computer Science, Peking University\\
    $^{3}$School of Information, Renmin University of China\hspace{0.3em}
    $^{4}$School of Integrated Circuit Science and Engineering, Beihang University
  }
  \country{}
}

\begin{document}

\begin{abstract}
To leverage user-specific data, retrieval-augmented generation is employed in multimodal LLM applications. 
While multi-vector retrieval (MVR) improve retrieval accuracy by query decomposition and image segmentation, they suffer from sub-optimal accuracy and efficiency due to the misalignment between sub-queries and varying image objects and redundant fine-grained image segments.
In this work, we present \textit{MIRAGE}, an efficient scheduling framework for image retrieval.
    First, we introduce a novel hierarchical paradigm, employing multiple intermediate granularities for varying image objects to enhance alignment.
    Second, we minimize the retrieval overhead by leveraging cross-hierarchy ranking consistency and hierarchy redundancy to eliminate unnecessary computation. 
    Furthermore, we configure parameters for each dataset automatically for practicality across diverse scenarios.
Experiments show that 
    \textit{MIRAGE} achieves substantial accuracy improvements and reduces computation by up to $3.5\times$ over the existing MVR system.
\end{abstract}

\maketitle

\begingroup
\renewcommand\thefootnote{}
\footnotetext{$^{*}$ Equal contribution.}
\footnotetext{$^{\dagger}$ Corresponding author: Xiang Chen, xiang.chen@pku.edu.cn.}
\endgroup



\section{\textbf{Introduction}}
\label{sec:introduction}

The rapid advancement of large language models (LLMs) has enabled emerging applications such as intelligent agents and personal assistants~\cite{li2024personal}.
	However, LLMs are not inherently capable of effectively leveraging user-specific data.
	To address this limitation, retrieval-augmented generation (RAG) has been widely adopted, where external data relevant to a user query is retrieved and incorporated into the generation process to improve accuracy and relevance~\cite{zhao2024ragsurvey}.   
Meanwhile, when RAG is extended to multimodal LLM systems, specifically with image retrieval, it computes the similarity between the feature vectors of language prompts and particular image objects~\cite{abootorabi2025askmodality}.
	Most multimodal RAG systems in production, however, follow a simple one-shot paradigm.
    As shown in Fig.~1, MVR embeds an entire query and an entire image into a single global vector, referred to as ``\textit{1 Mode}'' in this context.
	While efficient, this single-vector retrieval inevitably loses fine-grained object information, leading to unsatisfactory retrieval accuracy for complex or semantically diverse image content.

\begin{figure}[t]
    \centering
    \includegraphics[width=3.4in]{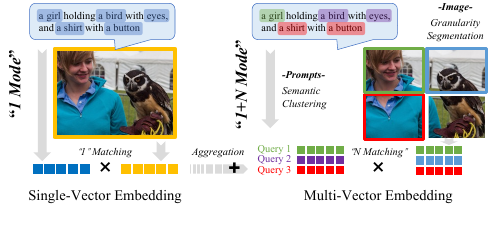}
    \vspace{-7mm}
    \caption{Multimodal Image Retrieval}
    \Description[A]{AA}
    \vspace{-5mm}
    \label{fig:fig-mvr-1-n}
\end{figure}

To address this, recent studies explore retrieval with data decomposition. As illustrated in Fig.~\ref{fig:fig-mvr-1-n}, multi-vector retrieval (MVR) decomposes a query into multiple sub-queries by semantic clustering with LLM prompts and decomposes an image into \textit{N} segments via granularity segmentation.
    These sub-query embeddings are matched against image segments, and then aggregated with the single-vector matching results to balance global image and fine-grained object granularity.
	Such ``\textit{1+N Mode}'' represents a fundamental shift toward better granularity of alignment between queries and the unpredictable semantic object composition of images.

Despite its accuracy benefits, MVR introduces several challenges in terms of efficiency and practicality.
	First, the number of decomposed vectors (\textit{N}) is often set to be large in order to capture fine-grained objects. Yet, selecting the optimal \textit{N} is highly non-trivial: small \textit{N} fails to fully represent image granularity, while large \textit{N} breaks the integrity of objects.
	Second, finer decomposition inherently amplifies computation complexity, as each additional image segment takes extra similarity calculation against multiple sub-queries.
	Finally, although finer-grained decomposition reveals structural properties such as information consistency across image segments and redundancy within query–image alignments, prior work has largely overlooked exploiting these opportunities~\cite{liu2025poqd,aiger2025globaltolocal}.

\begin{figure}[b]
    \centering
    \vspace{-4mm}
    \includegraphics[width=3.5in]{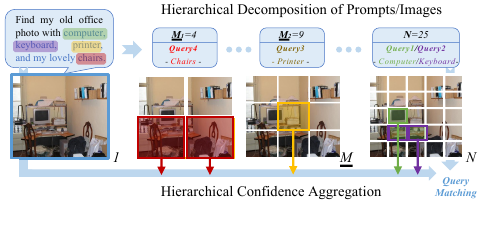}
    \vspace{-7mm}
    \caption{Image Retrieval with Hierarchical Decomposition}
    \Description[A]{AA}
    \vspace{-1mm}
    \label{fig:fig2}
\end{figure}

\textit{These challenges motivate the design of \textit{\textbf{MIRAGE}}, a runtime scheduling framework for multi-vector image retrieval:}
	\\\indent\textit{From the algorithm perspective}, unlike the conventional ``\textit{1+N Mode}'' MVR, \textit{\textbf{MIRAGE}} extends it into a ``\textit{1+\underline{M}+N} Mode'' as shown in Fig.~\ref{fig:fig2}, where \textit{\underline{M}} represents a series of intermediate image segmentation granularities and hierarchical query matching.
	This hierarchy effectively adapts to images with varying object scales, significantly improving alignment robustness and, therefore, retrieval accuracy.
    \\\indent\textit{From the computing perspective}, although ``\textit{1+\underline{M}+N Mode}'' seems to further increase the computation complexity compared to ``\textit{1+N}'' MVR, it actually exposes a more favorable computing optimization space.
	It exploits the cross-hierarchy consistency of retrieval information to reduce redundant matching, while detecting and pruning unnecessary hierarchies to minimize computational overhead.
	\\\indent\textit{From the design automation perspective}, \textit{\textbf{MIRAGE}} further enhances practicality by supporting automated configuration across datasets.
	Given the dataset variation, it conducts a lightweight profiling of the dataset and derives the optimal parameterization for hierarchical decomposition.
	This enables joint automation of algorithmic decomposition and computational scheduling, significantly improving adaptability across diverse deployment scenarios.

Following such a design methodology of \textit{\textbf{MIRAGE}}, this work made the following contributions:
    \textbf{(1)} We present a hierarchical decomposition framework for multi-vector image retrieval, offering a novel approach adaptive to diverse image granularities.
    \textbf{(2)} We systematically exploit redundancy in multi-vector retrieval, enabling runtime acceleration mechanisms that eliminate unnecessary computations.
    \textbf{(3)} We integrate these techniques into an automated framework that jointly optimizes accuracy and efficiency, enabling robust and adaptive deployment across scenarios.
    
To the best of our knowledge, this is the first hierarchical decomposition approach in the multimodal RAG domain.
    It offers an extensible algorithmic foundation that doubles the accuracy improvement of naive MVR, while achieving up to 3.5$\times$ computational cost reduction, approaching the single-vector retrieval efficiency.
\section{\textbf{Background}}
\label{sec:background}

\subsection{\textbf{Decomposition in Retrieval}}

In multimodal RAG systems, the retrieval task aims to identify the most relevant image for a given input prompt query.
    Both the query and images are embedded into a shared vector space using text–image embedding models~\cite{radford2021learning,li2023bge1}, and the top-$K$ images with the highest similarity scores to the query are retrieved.
    The retrieval process can be expressed as Eq.~\ref{equ:1}, where $Q$ denotes the query, $D$ the image set, and $E(\cdot)$ the embedding model.
    \textit{$\operatorname{SIM}$} represents a similarity operator (e.g., cosine similarity or dot product).
\begin{equation}
\label{equ:1}
    \underset{0 < i \le N_D}{\operatorname{TopK}}(\operatorname{Score}(Q, D_i)) = \operatorname{TopK}(\operatorname{SIM}(E(Q),E(D_i)))
\end{equation}

However, encoding a semantically complex query or image into a single vector inevitably loses information, resulting in sub-optimal accuracy.
To this end, one-shot decomposition-based methods, or MVR (e.g., ColBERT and its variants~\cite{khattab2020colbert,santhanam2022colbertv2}), have been proposed.
As formulated in Eq.~\ref{equ:MVR_N}, MVR decomposes a query into multiple sub-queries ${q_i}$ and each image into multiple segments ${D_{i,j}}$.
The overall score is computed as the product of the scores of all sub-queries, where the score of each sub-query is defined as the maximum similarity against all image segments.
Thus, each sub-query matches its most relevant image segment, ensuring that all semantic components of the query are satisfied.
To further improve retrieval accuracy, recent studies~\cite{liu2025poqd} adopt the ``\textit{1+N Mode}'', which aggregates scores computed with and without decomposition.

\vspace{-5pt}

\begin{equation}
\label{equ:MVR_N}
\operatorname{Score}(Q, D_i) = \operatorname{SIM}(E(Q), E(D_{i}))
     + \prod_{k=1}^{N_q} \max\limits_{j=1}^{N_g} \operatorname{SIM}(E(q_k), E(D_{i,j})).
\end{equation}

Despite of its accuracy benefits, there is still ample space for improving MVR: First, granularity selection and alignment in image decomposition has not been systematically explored. Second, it introduces heavy overhead as similarity calculation increases by tens of times due to ``\textit{N}'' matching.

\subsection{\textbf{Decomposition Granularity}}

The decomposition granularity in MVR, i.e., the number of decomposed vectors \textit{N}, plays a critical role in retrieval accuracy.  
Granularity can be considered from two perspectives: 

\noindent\textbf{Query decomposition granularity.}  
A query should be decomposed into semantically independent sub-queries to achieve precise alignment with image objects.
Coarse-grained decomposition often results in information loss, while overly fine-grained ones, like token-level~\cite{khattab2020colbert}, may break semantic integrity and cause spurious matches.  
Recent work~\cite{liu2025poqd} leverages fine-tuned LLMs~\cite{grattafiori2024llama3herdmodels} to adaptively decompose complex queries.

\noindent\textbf{Image decomposition granularity.}  
Unlike text retrieval, where segmentation boundaries naturally exist at the sentence or paragraph level, image retrieval lacks inherent structural boundaries.
Consequently, fast adaptive segmentation methods such as SLIC~\cite{achanta2012slic} are employed to partition images into segments. 
The major limitation of existing image decomposition approaches is that granularity is pre-defined and lacks runtime adaptability.  

\vspace{-4pt}
\subsection{\textbf{Approximate Retrieval Acceleration}}

Increasing retrieval scale makes computational redundancy non-trivial, driving the need for approximate methods that trade accuracy for efficiency.
    Most existing approximate acceleration techniques are algorithm-agnostic:  
    Indexing-based methods, such as IVF\cite{douze2024faiss}, HNSW~\cite{malkov2018approximate}, and PLAID~\cite{santhanam2022plaid}, reduce the number of retrieval candidates via clustering embedding vectors;  
    Quantization (e.g., IVF-PQ~\cite{douze2024faiss}) reduce similarity computation latency by lowering data bit-width, thereby enhancing hardware parallelism.  
    Meanwhile, our approach exploits redundancy inherent in the MVR algorithm itself, which is largely orthogonal to these techniques and can thus be seamlessly integrated into existing optimizations.   
\section{\textbf{Hierarchical Decomposition}}
\label{sec:decomp}

In this section, we analyze the misalignment between queries and image segments regarding retrieval granularities.
    We then propose a hierarchical decomposition framework, which enables better alignment and exposes new scheduling opportunities.

\subsection{\textbf{Granularity Misalignment Analysis}}
\label{sec:3.1}

\begin{figure}[t]
    \centering
    \includegraphics[width=3.3in]{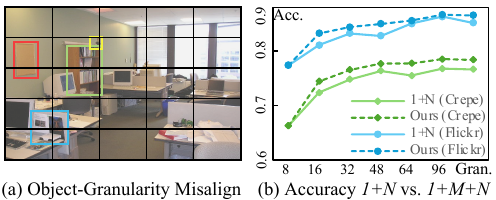}
    \vspace{-2mm}
    \caption{Decomposition Hierarchy and Accuracy Impact}
    \Description[A]{AA}
    \vspace{-2mm}
    \label{fig:fig-motivate-alg}
\end{figure}

While the ``\textit{1+N Mode}'' substantially improves accuracy, it still suffers from misalignment between fixed decomposition granularity and varying object scales.  
    As shown in Fig.~\ref{fig:fig-motivate-alg}(a), where the image is decomposed into 25 segments with ground-truth objects highlighted by colored rectangles, the misalignment becomes evident.  
    For certain objects, such as the corkboard (highlighted in red), decomposition is beneficial: irrelevant regions are cropped out while the main body of the object remains intact.
    However, other objects are adversely affected when the granularity is misaligned: objects may be fragmented into parts (e.g., the shelves marked in green) or merged with irrelevant regions (e.g., the toy marked in yellow).  

To further prove this observation, we conduct a preliminary experiment by profiling accuracy across a set of granularities using the ``\textit{1+N Mode}'' in Eq.~\ref{equ:MVR_N}.
The results on two datasets are plotted with solid lines in Fig.~\ref{fig:fig-motivate-alg}(b).
As the decomposition granularity becomes finer, the retrieval accuracy generally shows an upward trend and then levels off.
However, accuracy fluctuates at both certain granularities in the middle and the finest ones. This can be attributed to the misalignment.
Hence, to achieve higher retrieval accuracy, the decomposition granularity “\textit{N}’’ should ideally align with the scale of each object, which is infeasible with a single fixed granularity.
We therefore turn to the aggregation of multiple granularities.

\subsection{\textbf{Hierarchical Aggregation}}
Inspired by the analysis above, we want to align each object with its best-matching granularity to improve retrieval accuracy.
Thus, we extend the ``\textit{1+N Mode}'' by constructing the hierarchy of granularities and searching along it for the most suitable granularity of each sub-query.
The proposed algorithm can be denoted as ``\textit{1+\underline{M}+N Mode}'' as illustrated in Fig.~\ref{fig:fig2}.
Specifically, each image is hierarchically segmented into multiple granularities, each corresponding to a different patch size. We then iterate from coarse to fine granularities, computing the similarity score between each sub-query and image segments at every granularity as in Eq.~\ref{equ:MVR_N}. After each iteration, the score of each sub-query is updated by selecting the maximum similarity score among all examined granularities, representing its best alignment.
With granularity denoted $g$, segment count in granularity $g$ denoted $N_g$, and the total number of granularity hierarchies denoted $N_G$, we can extend Eq.~\ref{equ:MVR_N} into Eq.~\ref{equ:MVR_1MN}:

\vspace{-5pt}

\begin{equation}
\label{equ:MVR_1MN}
    \operatorname{Score}(Q, D_i)=\operatorname{SIM}(E(Q), E(D_{i})) 
    +\prod_{k=1}^{N_q}\max_{g=1}^{N_G} \max_{j=1}^{N_g} \operatorname{SIM}(E(q_{k}), E(D_{i,j}^g))
\end{equation}

Our ``\textit{1+\underline{M}+N} Mode'' is superior to the prevalent ``\textit{1+N Mode}'' in two aspects.
First, by considering all possible granularities, it adaptively leverages the most suitable segment size for each sub-query.
As shown in Fig.~\ref{fig:fig2}, the sub-query ``chairs'' matches the granularity of 4 while the sub-queries ``computer'' and ``keyboard'' match the granularity of 16. 
As shown in Fig.~\ref{fig:fig-motivate-alg}, the accuracy of our``\textit{1+\underline{M}+N} Mode''—which aggregates the scores of each sub-query across all prefix granularities—is always higher than ``\textit{1+N} Mode'' and grows monotonically as the granularity becomes finer.
Furthermore, our approach provides flexibility for an accuracy-performance trade-off by allowing for computing the relevance score on a selected subset of the hierarchy, which will be elaborated in the next section.
\section{\textbf{Computational Efficiency Exploration}}
\label{sec:query_process}

The proposed hierarchical framework decouples the retrieval process into structured levels, enabling systematic analysis of information redundancy and optimization of computational efficiency.

\begin{figure}[b]
    \centering
    \includegraphics[width=3.4in]{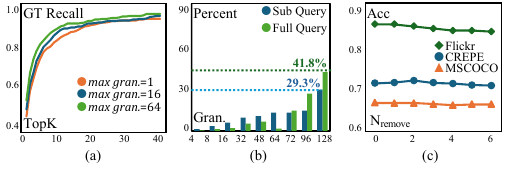}
    \vspace{-8mm}
    \caption{
        Redundancy Analysis:
        (a) Similarity of ground-truth image consistently ranks near the top across hierarchies.
        (b) The deepest hierarchies are not always needed.
        (c) Removing hollow hierarchies causes trivial accuracy loss.
    }
    \Description[A]{AA}
    \label{fig:fig-motivate-opt}
\end{figure}

\subsection{\textbf{Computational Redundancy Analysis}}

\begin{figure*}[t!]
    \centering
    \includegraphics[width=7.1in]{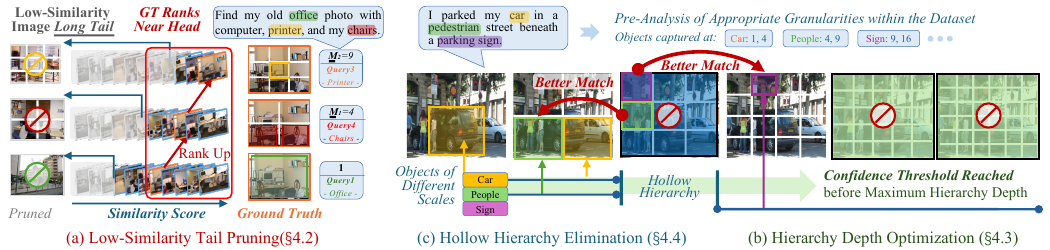}
    \vspace{-5mm}
    \caption{Computational Optimization in \textbf{\textit{MIRAGE}}}
    \label{fig:fig-overview}
    \Description[A]{AA}
    \vspace{-4mm}
\end{figure*}

Deriving from Eq.~\ref{equ:MVR_1MN}, computation redundancy can be attributed to two dimensions: image $D_i$ and granularity $D^g$.  

\textbf{Low-Similarity Image Long-Tail.}
Intuitively, coarse-grained image segments still preserve information about finer-grained objects, despite embedding loss.
We argue that the coarse-grained similarity scores obtained in the early iterations are already sufficient to distinguish relevant images from irrelevant ones.  
We validate this hypothesis by evaluating the top-$k$ recall rate of ground-truth at different hierarchy depths.
Fig.~\ref{fig:fig-motivate-opt}(a) shows that the ground-truth consistently ranks near the top. 
This suggests that further computing costly fine-grained similarity is redundant for a large portion of low-similarity images, which forms a long tail as shown in Fig.~\ref{fig:fig-overview}(a).

\textbf{Unnecessary Deep Traversal.}  
Although deeper hierarchies capture finer-grained objects, we observe that not all queries align with the finest-grained representations. According to Eq.~\ref{equ:MVR_1MN}, the similarity score between a query and an image stabilizes once all the sub-queries have found their highest-similarity patch.
Therefore, exhaustively traversing all hierarchies for each query introduces substantial redundancy.
To quantify this effect, we measure the distribution of (i) the granularity of the best-matching patch for each sub-query, and (ii) the deepest hierarchy at which the similarity score between the full query and an image converges, as shown in Fig.~\ref{fig:fig-motivate-opt}(b).
The results indicate that deep hierarchies are not always necessary for every query, offering optimization opportunities.

\textbf{Hollow Hierarchies.} When adjacent hierarchies differ only slightly, an object captured by an intermediate hierarchy is also likely to be captured by neighboring hierarchies. This results in redundant intermediate hierarchies, which are referred to as ``hollows''.
    Fig.~\ref{fig:fig-overview}(c) gives an example where both people and the sign are captured by granularity 9, yet higher-similarity matches exist at adjacent hierarchies.
    To evaluate this effect, we start from the granularity set with a stride of 8 and iteratively remove intermediate granularities until the final set has a stride of 16. 
    The results in Fig.~\ref{fig:fig-motivate-opt}(c) show that removing these redundancy hierarchies has minimal impact on accuracy.

Based on our analysis of computational redundancy, we identify three complementary dimensions of computational optimization within hierarchical decomposition.
As illustrated in Fig.~\ref{fig:fig-overview}, these dimensions highlight distinct opportunities for efficiency improvement, which we elaborate on in the following subsections.

\begin{algorithm}[b!]
\caption{Online Redundancy Reduction}
\label{alg:query_process}
\begin{algorithmic}[1]
\small

\Procedure{ProcessQuery}{$Q$, $\{q_k\}$}
  \State \textbf{Let} $\operatorname{Score}[N_D]$, $\operatorname{SubScore}[N_D, N_q]$, $\operatorname{TopK}[N_G]$
  \For{$g := 1$ to $N_G$}
    \For{$i := 1$ to $N_D\cdot t_{g-1}$}
      \For{$k := 1$ to $N_q$, $j := 1$ to $N_g$}
        \State $\operatorname{SubScore}_{i, k} := \max(\operatorname{SIM}(E(q_k), E(D_{i,j}^g)), \operatorname{SubScore}_{i,k})$
      \EndFor
      $\operatorname{Score}_{i} := \operatorname{SIM}(E(Q),E(D_i))+\prod _{k=1}^{N_q}\operatorname{SubScore}_{i, k}$
    \EndFor
    \State $D$ := $\operatorname{ArgSort}$($\operatorname{Score}$)[1:$N_D \cdot t_g$]
    \State $\operatorname{TopK}[g] := D[1:N_{K}]$
    \If{$\operatorname{K\tau}(\text{TopK}[g], \text{TopK}[g-1]) \geq \tau$}
      \State \textbf{break}
    \EndIf
  \EndFor
  \State \Return $\operatorname{TopK}[g]$
\EndProcedure
\end{algorithmic}
\end{algorithm}

\subsection{\textbf{Low-similarity Tail Pruning}}
\label{sec:image_topk}

Fig.~\ref{fig:fig-motivate-opt}(a) shows that the scores of ground-truth images consistently appear among the top-ranked images even with coarse granularities.  
    As finer granularities are introduced in deeper hierarchy levels, the top-ranked results become more precise; however, the cost of similarity computation increases substantially due to the expanded number of image segments. This motivates pruning low-ranked images from subsequent ranking updates to avoid redundant computation.
We propose an online optimization algorithm that tracks image rankings across iterations.
    By pruning expensive fine-grained evaluations for low-similarity candidates, the algorithm effectively reduces runtime overhead without sacrificing accuracy.

Specifically, we compute similarity scores for image segments in each granularity hierarchy and update the aggregated scores for images iteratively.
    In each iteration, images with low aggregated similarity scores are pruned and excluded from computation in subsequent hierarchies, as shown in Fig.~\ref{fig:fig-overview}(a).
    The procedure is elaborated in L.2--L.7 of Alg.~\ref{alg:query_process}.
    The ratio of images pruned in each granularity hierarchy is specified with $t_g=\alpha^g T$, where $T$ denotes the initial reduction ratio and $\alpha$ denotes the decay rate.
    Empirically, $T$ can be more aggressive, as ground truth usually ranks near the top. While it is better to set a conservative $\alpha$, in that the score rank of ground truth may fluctuate as the granularity becomes finer.
    

\subsection{\textbf{Hierarchy Depth Optimization}}
\label{sec:early_exit}

While Sec.~\ref{sec:image_topk} optimizes redundant fine-grained computation from the image side, we observe that similar opportunities also exist from the query side.
Depth distribution in Fig.~\ref{fig:fig-motivate-opt}(b) shows that a substantial portion of queries align with coarse-grained information, indicating that they do not require traversing the full hierarchy.
    For these queries, matching against the finest-grained segments introduces redundant computation.
    To address this, we develop an early-exit mechanism that monitors confidence as granularity becomes finer and exits once it exceeds a predefined threshold.

The key challenge is to define the confidence indicator.  
    A natural choice is to stop once the top-$k$ list stabilizes.  
    To quantify this convergence, we employ \textit{Kendall’s $\tau$ coefficient}~\cite{kendall1938new}, denoted $\operatorname{K\tau}$, which measures the ordinal association between two ranked sequences. As illustrated in Fig.~\ref{fig:fig-overview}(b), we compute $\operatorname{K\tau}$ for the top-$k$ rankings between consecutive iterations as a confidence measure and halt as soon as it reaches a certain threshold, based on the intuition that a stabilized $\operatorname{K\tau}$ indicates ranking convergence.
The procedure is detailed in Alg.~\ref{alg:query_process} (L.8--L.10), where $\textit{TopK}[g]$ denotes the set of $N_K$ images with the highest similarity scores up to granularity $g$.

\subsection{\textbf{Hollow Hierarchy Elimination}}

Hollow hierarchy analysis further indicates that some intermediate hierarchies are unnecessary as they substantially overlap with their neighbors.
However, making skip choices online is impractical without any prior knowledge, and pre-defining the hierarchical granularity set lacks adaptability.
Therefore, we treat the granularity set as a tunable parameter and propose an offline hierarchy elimination algorithm to remove redundant hierarchies. Specifically, we run the algorithm on a validation query split while the image database remains unchanged, allowing us to optimize the hierarchy based on performance observed on this split.

Alg.~\ref{alg:alg2} (L.1--L.12) details the algorithm. 
    We construct the hierarchy by initializing granularities spaced at intervals of $S_G$, where a smaller $S_G$ yields higher flexibility but more redundancy. 
    The algorithm iteratively removes the granularity with the least accuracy loss before reaching a threshold.
    To ensure scheduling flexibility, consecutive removals are prohibited to prevent scenarios where only fine-grained hierarchies remain, limiting opportunities for online optimizations.
Fig.~\ref{fig:fig-overview}(c) shows an example of this approach, where the granularity of 9 is considered non-essential and removed from the hierarchy, thereby reducing computational overhead.


\section{\textbf{Framework Integration \& Automation}}
\label{sec:impl}

\begin{algorithm}[t]
\caption{Automated Configuration}
\label{alg:alg2}
\small
\begin{algorithmic}[1]


\Procedure{SetGran}{$S_G, \tau, T, \alpha$}
  \State $G := 0$; $\{N_g\} := \emptyset$
  \Repeat
    \State $G := G+S_G;\ \{N_g\} := \{N_g\} \cup G$
    \Comment{Init hierarchy with stride $S_G$}
  \Until{$\operatorname{EvalAcc}(\{N_g\},  \tau, T, \alpha)$ converge to $\mathcal{A}$}
  \Repeat
    \For{$i := 1$ to $|\{N_g\}|$}
      \State $Acc_i := \operatorname{EvalAcc}(\{N_g\}-N_i, \tau, T, \alpha)$
    \EndFor
    
    \State $g_\text{remove} := \operatorname{argmax}(\{Acc_i \mid N_i $ not near the last removed one$\})$
    \State $\{N_g\} := \{N_g\} - N_{g_\text{remove}}$
  \Until{$\operatorname{EvalAcc}(\{N_g\}, \tau, T, \alpha) \leq \mathcal{A}-\epsilon$}
  \State \Return $\{N_g\}$
\EndProcedure

\Procedure{Configure}{$R_L, R_S, R_\tau, R_T, R_\alpha$}
  \For{$L, S, \tau, T, \alpha$ in $R_L, R_S, R_\tau, R_T, R_\alpha$}
      \State Let $\{N_g\} := \Call{SetGran}{S, \tau, T, \alpha}$
      \If{$L \geq \operatorname{Latency}(\{N_g\}, \tau, T, \alpha)$ \textbf{and} $\operatorname{EvalAcc}(\text{Cfg}[L]) < \operatorname{EvalAcc}(\{N_g\}, \tau, T, \alpha)$}
        \State $\text{Cfg}[L] := \{\{N_g\}, \tau, T, \alpha\}$
      \EndIf
  \EndFor
\EndProcedure

\end{algorithmic}
\end{algorithm}

\subsection{\textbf{Automated Configuration}}
Based on the redundancy theory introduced in Section~\ref{sec:query_process}, we implement \textbf{\textit{MIRAGE}} with an automated configuration algorithm for optimizing parameters for various trade-off objectives. 

\textbf{\textit{MIRAGE}} can flexibly adapt to different dataset and deployment scenarios, whether accuracy or performance oriented, by exposing a set of tunable parameters:  
    the initial ratio $T$ and decay rate $\alpha$ for low-similarity image pruning,  
    the hierarchy early-exit threshold $\tau$,  
    and the granularity initialization stride $S_G$.

Efficiently optimizing over this high-dimensional parameter space is non-trivial.  
    Thus, we design a latency-guided configuration algorithm based on grid search, as presented in Alg.~\ref{alg:alg2}.  
    For each parameter, the search range and step size are predefined.  
During optimization, we traverse the latency dimension to limit profiling overhead, guided by a lightweight performance model:  
\begin{equation}
\label{equ:perf_model}
    \operatorname{Latency} = N_q \sum\nolimits_{g=1}^{N_G} N_g \times t_g \times N_D,
\end{equation}
\noindent where  
$N_D$ denotes the image set size,  
$N_G$, $N_q$, $N_g$ are the granularities, sub-queries, and the segments per granularity $g$,  
and $t_g$ is the fraction of images preserved at hierarchy $g$.  

Since similarity computation is lightweight, the latency of matching each sub-query with an image segment can be treated as a constant and is omitted for simplicity.  
Constrained by latency, we first establish the granularity hierarchy, which requires more extensive exploration (L.15). 
    Subsequently, other parameters are tuned for the highest accuracy under each latency constraint as (L.16--L.17).  
    The latency-guided search enables efficient automated configuration.  

It is worth noting that, in most retrieval scenarios~\cite{dubey2021retrievalsurvey}, the database remains relatively static and can therefore be pre-analyzed offline to improve retrieval efficiency. 
    We adopt the same assumption and use a separate validation query split solely for parameter tuning. 
    With proper initialization, the entire preprocessing procedure finishes within tens of minutes on the datasets used in our experiments, which is acceptable for practical deployment.

\vspace{-2mm}
\subsection{\textbf{Implementation Detail}}

We added about 1.1k lines of Python code based on the open-source code from ~\cite{liu2025poqd} built on PyTorch~\cite{pytorch} and Faiss~\cite{douze2024faiss}, implementing our proposed scheduling framework and automated configuration algorithm, where the latency is profiled with CUDA Event API.
    For image decomposition, given that the boundaries of segments generated by SLIC are irregular, we pad the remaining area with a black background to form rectangular patches.
    For query decomposition, we deploy a local large language model using vLLM~\cite{kwon2023efficient}, and the prompt template is adopted directly from \cite{liu2025poqd} for fair comparison.
\section{\textbf{Experiments}}
\label{sec:experiments}

\begin{table*}[t!]
\vspace{-12pt}
\caption{Main Results}
\vspace{-4mm}
\label{tab:main_result}
\centering
\footnotesize
\resizebox{\textwidth}{!}{%
\begin{tabular}{ccccccccccc}
\toprule
\multirow{2}{*}{Datasets} &
  \multicolumn{2}{c}{CREPE} &
  \multicolumn{2}{c}{MSCOCO} &
  \multicolumn{2}{c}{NoCaps} &
  \multicolumn{2}{c}{Flickr} &
  \multicolumn{2}{c}{Improvement(Avg.)} \\ \cmidrule{2-11} 
                  & NDCG@10 & QPS   & NDCG@10 & QPS   & NDCG@10 & QPS & NDCG@10  & QPS   & Spd.  & Acc.                     \\ \midrule
Vanilla           & 65.11   & 1230  & 66.27   & 678.1 & 82.10   & 638.4 & 82.30  & 601.4 & ---   & ---                      \\
POQD              & 71.62   & 54.8  & 68.44   & 27.8  & 84.68   & 27.8  & 85.19  & 29.7  & 1x    & 1x(+3.53)                \\
Ours (w/o Opt.)   & 73.22   & 12.1  & 70.13   & 6.7   & 85.60   & 7.1   & 87.70  & 6.9   & 0.25x & 1.48x(+5.22)             \\
Ours (w/o O1,O2)  & 73.17   & 21.1  & 70.39   & 10.8  & 85.45   & 10.9  & 87.59  & 11.3  & 0.4x  & 1.47x(+5.2)              \\
Ours (w/o O1)     & 73.56   & 133.3 & 70.60   & 78.1  & 85.84   & 76.23 & 87.24  & 72.8  & 2.5x  & \textbf{1.52x(+5.37)}    \\
Ours              & 72.83   & 148.6 & 70.27   & 93.3  & 85.23   & 101.3 & 86.57  & 102.9 & \textbf{3.5x}  &  1.43x(+5.03)   \\ \bottomrule
\end{tabular}%
}

\end{table*}

\subsection{\textbf{Experiment Setup}}
\noindent\textbf{Evaluation Platform}. 
We evaluate our framework on an edge server-level platform equipped with an Intel Xeon 4410T CPU and one NVIDIA A100-PCIE-40G GPU.

\noindent\textbf{Baselines}.
We compare our framework against two baselines:
    (1) \textit{Vanilla dense retrieval}: the conventional approach that encodes each query and data item as a single vector, corresponding to the ``\textit{1 Mode}'' in Eq.~\ref{equ:1}.
    (2) \textit{POQD}~\cite{liu2025poqd}: a state-of-the-art multi-vector retrieval framework matching decomposed query with image segments, corresponding to the ``\textit{1+N Mode}'' in Eq.~\ref{equ:MVR_N}.
Note that the proposed techniques are designed specifically for retrieval acceleration. Therefore, we do not include the LLM-based query decomposing process in our performance comparison.

\noindent\textbf{Datasets}. 
Extending the setup of \cite{liu2025poqd}, we evaluate our method and baselines on 4 text-image retrieval datasets: CREPE~\cite{ma2023crepe}, MSCOCO~\cite{lin2014mscoco}, NoCaps~\cite{agrawal2019nocaps}, and Flickr~\cite{young2014image}, with 1K, 40K, 2K, and 2K images, respectively, after data cleaning.
Considering the scale of MSCOCO, a subset is randomly sampled for experiments.
Note that we treat each image caption as a query and consider a caption relevant only to its corresponding image, following \cite{ma2023crepe}. A subset of the captions is then selected as the test query set.

\noindent\textbf{Models.}
Embedding and decomposition models are critical to retrieval accuracy.
For text-image aligned embedding, we employ CLIP~\cite{radford2021learning} to embed both queries and images.
For query decomposition, we use Qwen3-8B~\cite{yang2025qwen3} in place of the model used in \cite{liu2025poqd}, owing to its improved decomposition performance.

\noindent\textbf{Metrics.}
To evaluate efficiency, we measure the query throughput (queries per second, QPS) of \textbf{\textit{MIRAGE}} and baselines.
For retrieval accuracy, we report the Normalized Discounted Cumulative Gain (NDCG~\cite{wang2013ndcg}) at top-10, denoted as NDCG@10, following \cite{liu2025poqd}.

\subsection{\textbf{End-to-End Results}}
The main experimental results are summarized in Tab.~\ref{tab:main_result}.  
    We set four configurations to demonstrate the effectiveness of each proposed technique:  
    (1) only ``\textit{1+\underline{M}+N}'' algorithm without redundancy reduction. 
    (2) hollow hierarchy elimination (\textit{O3}),  
    (3) low-similarity tail pruning (\textit{O2}), and  
    (4) hierarchy depth optimization (\textit{O1}).  
    The parameter setting is obtained through the automated configuration algorithm.  
    Note that we mainly compare our framework with MVR systems (POQD baseline), and thus, the average speedup and accuracy improvement of the Vanilla baseline are omitted.
    
\begin{figure}[t]
    \centering
    \includegraphics[width=3.4in]{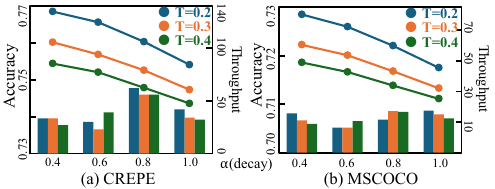}
    \vspace{-8mm}
    \caption{Trade-off Analysis: Low-Similarity Tail Pruning}
    \vspace{-4mm}
    \label{fig:exp-obligation-1}
    \Description[A]{AA}
\end{figure}

\textbf{\textit{MIRAGE}} achieves the best balance between accuracy and performance.
\textbf{\textit{MIRAGE}} achieves the highest accuracy with \textit{O3} and \textit{O2} applied, surpassing Vanilla and POQD by up to 8 and 2 percentage points.
Despite the accuracy loss with redundancy reduction, MIRAGE still beats the others.
In terms of performance, \textbf{\textit{MIRAGE}} is only second to Vanilla 
and achieves up to $4\times$ speedup than POQD.  
Across all \textbf{\textit{MIRAGE}} configurations, the plain hierarchical algorithm suffers most from excessive redundancy, though significant accuracy improvement is attained over POQD.
Introducing \textit{O3} doubles throughput with negligible accuracy loss.
Enabling \textit{O2} further boosts throughput by nearly 7$\times$, and surprisingly achieves the best accuracy.
Applying \textit{O1} yields an additional 40\% improvement.  

\subsection{\textbf{Analysis}}

\begin{figure}[t]
    \centering
    \includegraphics[width=3.4in]{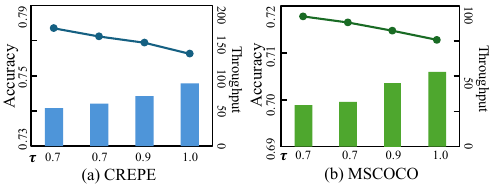}
    \vspace{-8mm}
    \caption{Trade-off Analysis: Hierarchy Depth Opt.}
    \vspace{-4mm}
    \label{fig:exp-obligation-2}
    \Description[A]{AA}
\end{figure}

Fig.~\ref{fig:exp-obligation-1} analyzes low-similarity tail pruning.  
    The curve shows throughput while the bar shows accuracy.  
    As expected, filtering more images at each hierarchy (i.e., reducing $T$ and $\alpha$) boosts throughput.  
    However, computation does not always translate to accuracy.  
    For example, setting $\alpha$=0.8 yields the highest accuracy on CREPE, surpassing $\alpha$=1.0.  
    It is common in CREPE that some images share common fine-grained objects but differ in composition.
    Thus, filtering tail images helps eliminate such interference.  
    This phenomenon is less evident in MSCOCO, since CREPE has higher visual complexity.
Fig.~\ref{fig:exp-obligation-2} investigates hierarchy depth optimization.  
    The trend is consistent across datasets: a larger early-exit threshold $\tau$ improves accuracy by considering more granularities for finer query–image alignment, at the expense of throughput.  
    This again highlights the intuitive performance–accuracy trade-off.

\begin{figure}[b]
    \centering
    \vspace{-4mm}
    \includegraphics[width=3.4in]{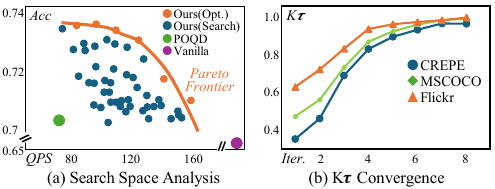}
    \vspace{-6mm}
    \caption{Component Analysis}
    \label{fig:exp-obligation-3}
    \Description[A]{AA}
\end{figure}

Although more computation generally favors accuracy, the trade-off between accuracy and performance is non-trivial, thus necessitating automatic configuration.  
    As shown in Fig.~\ref{fig:exp-obligation-3}(a), \textbf{\textit{MIRAGE}} consistently pushes the Pareto frontier.  
    The bottom-left green dot represents POQD, while Vanilla lies outside the figure due to its extremely high throughput but poor accuracy.
    The blue dots represent the samples generated in our search space, and among these marked with orange color are candidates reaching the optimal trade-off. 
 
Fig.~\ref{fig:exp-obligation-3}(b) illustrates the evolution of \textit{Kendall’s $\tau$ coefficient} over iterations. The consistent convergence of $\operatorname{K\tau}$ across datasets confirms its reliability as an indicator for early exit.

The scheduling overhead of \textbf{\textit{MIRAGE}} consists of two parts.  
    The first is the runtime early-exit metric computation, which compares the ranking of top candidates across consecutive iterations.  
    This step has constant time complexity.  
    The second is the additional sorting per iteration.  
    Since the number of granularities is small, the overall overhead is negligible (\textless 0.1\,ms) per query.  

\section{\textbf{Conclusion}}
\label{sec:conclusion}


We introduced \textbf{\textit{MIRAGE}}, a hierarchical decomposition framework for multi-vector image retrieval. By extending conventional ``\textit{1+N} Mode'' retrieval into ``\textit{1+\underline{M}+N} Mode,'' \textbf{\textit{MIRAGE}} adapts to diverse image granularities while systematically exploiting computational redundancy to improve efficiency. Through this co-design of algorithm, computation, and automation, \textbf{\textit{MIRAGE}} achieves significant gains in both accuracy and runtime cost, making fine-grained multimodal retrieval practical for real-world deployment. \textbf{\textit{MIRAGE}} opens up opportunities for broader integration and optimization with multimodal LLM systems and specific applications.

\balance
\clearpage
\bibliographystyle{ACM-Reference-Format}
\bibliography{
    _ref/1_MVR,
    _ref/2_dataset,
    _ref/3_related,
    _ref/98_framework_tool
}

\end{document}